\documentclass[11pt,a4paper]{article}
\usepackage[hyperref]{ranlp2025}
\usepackage{times}
\usepackage{latexsym}

\usepackage{microtype}

\usepackage{balance}

\usepackage[T1]{fontenc}
\usepackage{graphicx}
\usepackage{booktabs}
\usepackage{multirow}
\usepackage{graphicx}
\usepackage[normalem]{ulem}
\useunder{\uline}{\ul}{}
\usepackage{adjustbox} 
\usepackage{array} 
\usepackage[misc]{ifsym}

\usepackage{mwe}
\usepackage{url}

\aclfinalcopy

\title{The Illusion of a Perfect Metric: Why Evaluating AI's Words Is Harder Than It Looks}

\author{Maria Paz Oliva \\
  Iris AI  \\
  Borgen, Norway \\\And
  Adriana Correia \\
  Iris AI  \\
  Borgen, Norway \\\And
   Ivan Vankov\\
  Iris AI  \\ 
  Neurobiology BAS \\
  Sofia, Bulgaria \\\And
  Viktor Botev \\
  Iris AI  \\
  Sofia, Bulgaria
  }

\date{}

\begin{document}
\maketitle


\author{}
\author{Maria Paz Oliva\inst{1}\orcidID{0009-0009-6409-2083} \and
Adriana Correia\inst{1}\orcidID{0000-0003-0008-9167} \and
Ivan Vankov\inst{2,3}\orcidID{0000-0002-2772-5365} \and Viktor Botev\inst{2}\orcidID{0009-0004-6596-1694}}




\begin{abstract} Evaluating Natural Language Generation (NLG) is crucial for the practical adoption of AI, but has been a longstanding research challenge. While human evaluation is considered the de-facto standard, it is expensive and lacks scalability. Practical applications have driven the development of various automatic evaluation metrics (AEM), designed to compare the model output with human-written references, generating a score which approximates human judgment. Over time, AEMs have evolved from simple lexical comparisons, to semantic similarity models and, more recently, to LLM-based evaluators. However, it seems that no single metric has emerged as a definitive solution, resulting in studies using different ones without fully considering the implications. This paper aims to show this by conducting a thorough examination of the methodologies of existing metrics, their documented strengths and limitations, validation methods, and correlations with human judgment. We identify several key challenges: metrics often capture only specific aspects of text quality, their effectiveness varies by task and dataset, validation practices remain unstructured, and correlations with human judgment are inconsistent. Importantly, we find that these challenges persist in the most recent type of metric, LLM-as-a-Judge, as well as in the evaluation of Retrieval Augmented Generation (RAG), an increasingly relevant task in academia and industry.
Our findings challenge the quest for the ‘perfect metric’. We propose selecting metrics based on task-specific needs and leveraging complementary evaluations and advocate that new metrics should focus on enhanced validation methodologies.
\end{abstract}

\balance
\begin{figure*}[h]
    \centering
    \makebox[\textwidth]{\includegraphics[width=\linewidth]{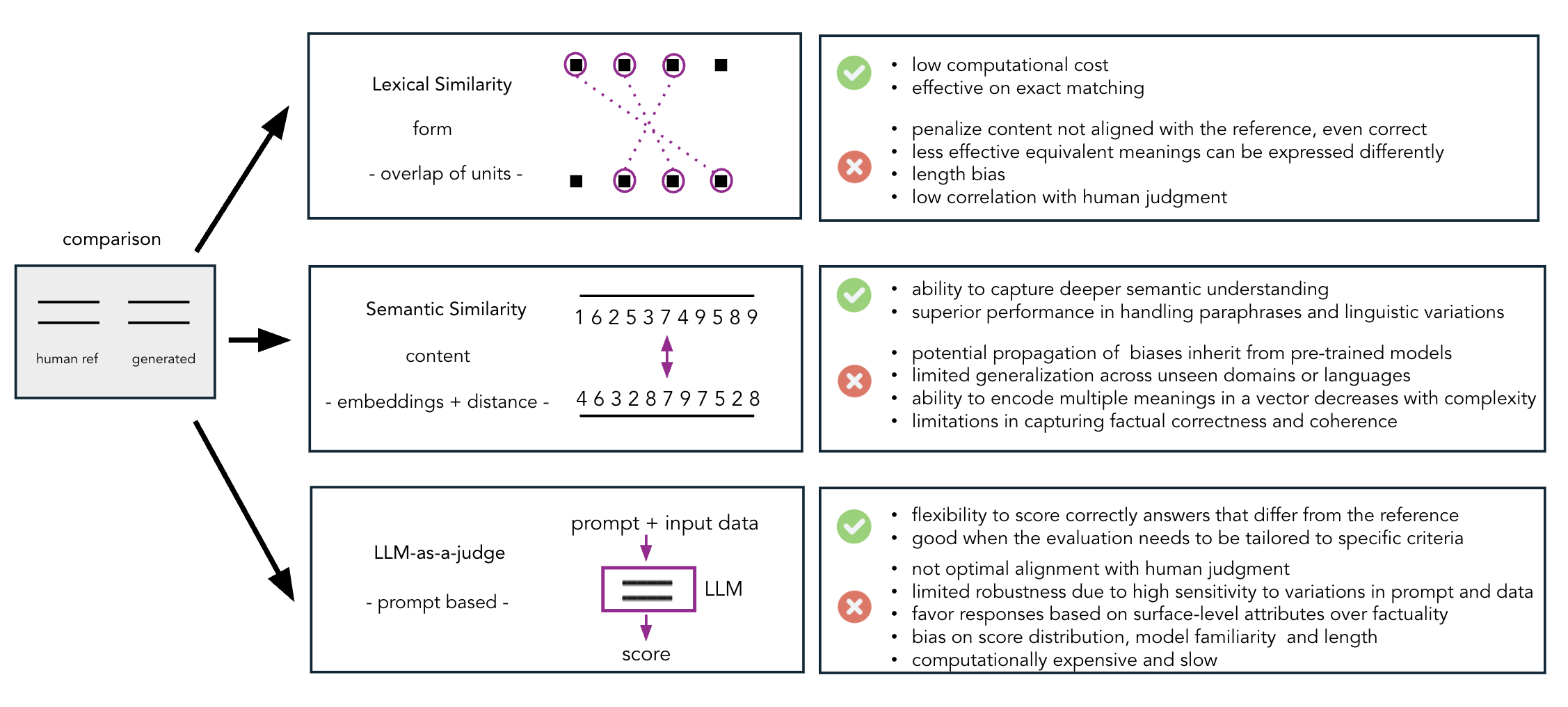}}
    \caption{Overview of evaluation metric categories, including lexical similarity, semantic similarity, and LLM-based evaluation, highlighting their advantages and limitations.}
    \label{fig:fig1}
\end{figure*}

\section{Introduction}
Evaluation of Natural Language Generation (NLG) models is essential to ensure their quality and usefulness in various real-world applications \cite{celikyilmaz2020evaluation}. As AI adoption expands, the importance of robust evaluation methods increases significantly \cite{novikova2017we}. However, assessing NLG outputs remains a challenging task \cite{gehrmann2023repairing}.
The standard approach to date is evaluating models' outputs through human judgment, but due to the high costs and limited scalability, the development of Automated Evaluation Metrics (AEM) has been pursued as an alternative \cite{clark2021all}. Traditionally, the supervised approach involves comparing model-generated text with a human-written reference. It assigns a similarity score based on the assumption that a higher similarity indicates a better quality of the output. Furthermore, because the score is considered meaningful, it is expected to serve as a good proxy for human judgment. This reliability is typically assessed by measuring the correlation between metric scores and human evaluations of the generated texts \cite{belz2006comparing,coughlin2003correlating}.

Evaluation metrics have been developed for decades while, simultaneously, demands have increased. Model architectures have become more advanced and complex evaluation methods have improved and become more sophisticated \cite{zhuang2023through}. The first evaluation scores were based on a comparison of the lexical surface forms of the generated text and a reference. Later, the introduction of embeddings enabled promising semantic comparisons, and more recently large language models (LLMs) have become the state-of-the-art in AEMs \cite{chang2024survey}.

With the wide range of available metrics, there seems to be a little agreement in the field about which ones are most effective and for what use cases. This has led to cases where multiple metrics are applied in studies without sufficient reflection on their implications and meaning \cite{mathur2020tangled}. An often overlooked fact is that most metrics were originally designed for specific NLP tasks, such as machine translation (MT), summarization, and question answering, raising concerns about their applicability to each other \cite{liguori2023who}. Furthermore, as we will demonstrate, inconsistencies arise when comparing the behavior of these metrics and their correlation with human judgment \cite{laskar2024systematic}.

The purpose of this article is to conduct an in-depth analysis of existing metrics. First, we analyze their methodologies, and then their documented strengths and limitations. Our focus is on how the metrics have been validated and their correlation with human judgment. In this work, we claim that the presented correlation is usually not sufficiently robust to justify the use of the metrics in a general use case scenario. 

The paper is structured as follows: Section \ref{metrics_analysis} presents existing evaluation metrics, analyzes their correlation with human judgment, and looks deeper into Retrieval Augmented Generation (RAG) as a real-world system evaluation example; Section \ref{threats_validity} discusses potential threats to validity; and Section \ref{conclusion} concludes the argument about the need for more thorough analysis of what aspects each metric evaluates and what synergies between metrics are necessary to have a reliable general framework of AEMs for model-generated text.

\section{Metrics Analysis} \label{metrics_analysis}

\subsection{Metrics Overview} \label{metrics_overview}
Based on their underlying scoring methodologies, evaluation metrics can be categorized into three groups: \textit{lexical similarity}, which measures the overlap of textual units; \textit{semantic similarity}, which quantifies the numerical distance between content representations; and \textit{LLM-based} evaluation, which relies on direct prompting of LLM for judgment \cite{10643089}. Figure \ref{fig:fig1} presents an overview of the main computational characteristics of each metric. An often overlooked aspect of these metrics is the initial context in which they were introduced, which typically involves the reporting of good correlations with human judgments. In Table \ref{tab:metrics_table}, we compare the different tasks, datasets, level, and metrics that were used to this date. In general, all works report good correlations, but we see that they are not directly comparable, so it is not obvious that one metric would perform better under the conditions of another one. In this section, we present an overview of key metrics within each category: examining their main characteristics, advantages, and limitations; as well as challenges to the initial claims of good correlation with human judgment.

\begin{table*}
\centering
\rotatebox{90}{\resizebox{\textheight}{!}{%
\begin{tabular}{|l|c|c|c|l|lll|}
\hline
\multirow{2}{*}{\textbf{Category}} &
  \multirow{2}{*}{\textbf{Metric}} &
  \multirow{2}{*}{\textbf{Year}} &
  \multirow{2}{*}{\textbf{Task}} &
  \multicolumn{1}{c|}{\multirow{2}{*}{\textbf{Distinctive Features}}} &
  \multicolumn{3}{c|}{\textbf{Initial Validation}} \\ \cline{6-8} 
 &
   &
   &
   &
  \multicolumn{1}{c|}{} &
  \multicolumn{1}{c|}{\textbf{Datasets}} &
  \multicolumn{1}{c|}{\textbf{Correlation Metrics}} &
  \multicolumn{1}{c|}{\textbf{Level}} \\ \hline
\multirow{3}{*}{\textbf{\begin{tabular}[c]{@{}l@{}}Lexical \\ Similarity\end{tabular}}} &
  \textbf{BLEU} &
  2002 &
  MT &
  \begin{tabular}[c]{@{}l@{}}Clipping (avoids over-counting), \\ brevity penalty, geometric mean\end{tabular} &
  \multicolumn{1}{l|}{\begin{tabular}[c]{@{}l@{}}Chinese-English, \\ monolingual\end{tabular}} &
  \multicolumn{1}{l|}{Pearson correlation} &
  system \\ \cline{2-8} 
 &
  \textbf{ROUGE} &
  2004 &
  Summarization &
  \begin{tabular}[c]{@{}l@{}}Variants: n-gram overlap, \\ LCS, skip-bigrams, etc.\end{tabular} &
  \multicolumn{1}{l|}{DUC 2001-2003} &
  \multicolumn{1}{l|}{\begin{tabular}[c]{@{}l@{}}Pearson, Spearman, \\ Kendall correlations\end{tabular}} &
  system \\ \cline{2-8} 
 &
  \textbf{METEOR} &
  2007 &
  MT &
  \begin{tabular}[c]{@{}l@{}}Alignment-based, considers \\ synonyms, stemming, paraphrases\end{tabular} &
  \multicolumn{1}{l|}{\begin{tabular}[c]{@{}l@{}}LDC TIDES 2003 \\ Arabic-to-English, \\ Chinese-to-English\end{tabular}} &
  \multicolumn{1}{l|}{Pearson correlation} &
  \begin{tabular}[c]{@{}l@{}}system, \\ segment\end{tabular} \\ \hline
\multirow{5}{*}{\textbf{\begin{tabular}[c]{@{}l@{}}Semantic\\ Similarity\end{tabular}}} &
  \textbf{WMD/SMS} &
  \begin{tabular}[c]{@{}c@{}}2015 / \\ 2019\end{tabular} &
  \begin{tabular}[c]{@{}c@{}}WMD: \\ Classification \\ SMS: \\ Summarization \\ and Essay\end{tabular} &
  \begin{tabular}[c]{@{}l@{}}Measures word/sentence \\ movement using word2vec\end{tabular} &
  \multicolumn{1}{l|}{\begin{tabular}[c]{@{}l@{}}WMD: BBCSPORT, \\ TWITTER, and others \\ SMS: CNN/Daily Mail news, \\ Hewlett Foundation’s essays\end{tabular}} &
  \multicolumn{1}{l|}{\begin{tabular}[c]{@{}l@{}}WMD: kNN error rate \\ SMD: Spearman \\ correlation\end{tabular}} &
  \begin{tabular}[c]{@{}l@{}}WMD: -\\ SMS: system\end{tabular} \\ \cline{2-8} 
 &
  \textbf{WISDM} &
  2017 &
  \begin{tabular}[c]{@{}c@{}}Classification, \\ Topic Prediction,\\ Filtering\end{tabular} &
  \begin{tabular}[c]{@{}l@{}}Uses TF-IDF and RV \\ coefficient for efficient similarity\end{tabular} &
  \multicolumn{1}{l|}{CORE database} &
  \multicolumn{1}{l|}{-} &
  - \\ \cline{2-8} 
 &
  \textbf{BERTScore} &
  2020 &
  \begin{tabular}[c]{@{}c@{}}MT, captioning, \\ paraphrase\end{tabular} &
  \begin{tabular}[c]{@{}l@{}}Uses BERT embeddings, \\ cosine similarity\end{tabular} &
  \multicolumn{1}{l|}{\begin{tabular}[c]{@{}l@{}}WMT 2016-2018,\\ IWSLT14 German-to-English, \\ COCO 2015, QQP, PAWS\end{tabular}} &
  \multicolumn{1}{l|}{\begin{tabular}[c]{@{}l@{}}Pearson, \\ Kendall correlations\end{tabular}} &
  \begin{tabular}[c]{@{}l@{}}system, \\ segment\end{tabular} \\ \hline
\multirow{5}{*}{\textbf{\begin{tabular}[c]{@{}l@{}}LLM-as\\ -a-judge\end{tabular}}} &
  \textbf{BARTScore} &
  2021 &
  \begin{tabular}[c]{@{}c@{}}Summarization, \\ MT, D2T\end{tabular} &
  \begin{tabular}[c]{@{}l@{}}Uses BART autoencoder, \\ computes log probability\end{tabular} &
  \multicolumn{1}{l|}{\begin{tabular}[c]{@{}l@{}}WMT19, REALSumm, \\ SummEval, \\ NeR18, QAGS\end{tabular}} &
  \multicolumn{1}{l|}{\begin{tabular}[c]{@{}l@{}}Spearman, \\ Kendall correlations\end{tabular}} &
  \begin{tabular}[c]{@{}l@{}}system, \\ dataset, \\ segment\end{tabular} \\ \cline{2-8} 
 &
  \textbf{Prometheus} &
  2023 &
  Long-form NLG &
  \begin{tabular}[c]{@{}l@{}}Fine-tuned LLAMA \\ for rubric-based scoring\end{tabular} &
  \multicolumn{1}{l|}{\begin{tabular}[c]{@{}l@{}}Feedback Bench, \\ Vicuna Bench\end{tabular}} &
  \multicolumn{1}{l|}{Pearson correlation} &
  segment \\ \cline{2-8} 
 &
  \textbf{AlpacaEval LC} &
  2023 &
  LLM evaluation &
  \begin{tabular}[c]{@{}l@{}}GPT-4 with length control \\ (LC) to prevent bias\end{tabular} &
  \multicolumn{1}{l|}{Chatbot Arena} &
  \multicolumn{1}{l|}{Spearman correlation} &
  system \\ \cline{2-8} 
 &
  \textbf{GPTScore} &
  2024 &
  \begin{tabular}[c]{@{}c@{}}Summarization, \\ MT,  dialogue\end{tabular} &
  \begin{tabular}[c]{@{}l@{}}Uses multiple LLMs for fluency, \\ coherence, informativeness\end{tabular} &
  \multicolumn{1}{l|}{\begin{tabular}[c]{@{}l@{}}FED, SummEval, \\ REALSumm, NEWSROOM \\ QAGS\_XSUM, BAGEL, \\ SFRES, MQM-2020\end{tabular}} &
  \multicolumn{1}{l|}{\begin{tabular}[c]{@{}l@{}}Pearson, \\ Spearman correlations\end{tabular}} &
  \begin{tabular}[c]{@{}l@{}}dataset, \\ segment\end{tabular} \\ \hline
\end{tabular}%
}
}
\caption{Overview of initial introduction of metrics.}
\label{tab:metrics_table}
\end{table*}

\subsubsection{Lexical similarity metrics}
These metrics focus on the surface-level form of the generated and reference texts, and the exact co-occurrence of sequences, such as n-grams. Sequences can be words, characters, strings, or tokens. Lexical metrics are considered as the traditional approach to evaluation in NLG ("traditional metrics"). Some relevant examples of this type of metrics are: \textbf{BLEU} \cite{papineni2002bleu}, \textbf{ROUGE} \cite{lin2004rouge} and \textbf{METEOR} \cite{banerjee2005meteor} (see Table \ref{tab:metrics_table}).

Lexical similarity metrics are widely used because of their ease of implementation and computational efficiency. These metrics are especially useful when the generated text must closely match the reference or when the output for evaluation is short. Nonetheless, lexical metrics have limitations in contexts where the generated text diverges from the reference but still conveys accurate information \cite{liu2016not}. They may yield a low score despite the identical meaning (paraphrased use case), which questions their correlation with human assessments \cite{coughlin2003correlating}. Furthermore, lexical metrics are less effective for complex texts with varied lexical and syntactic choices. They also suffer from length bias, which reduces their reliability for nuanced tasks such as quality in question generation \cite{nema2018towards} and dialogue response generation \cite{liu2016not}.

In fact, lexical-based metrics have struggled to demonstrate a strong correlation with human judgment since their introduction \cite{coughlin2003correlating}. Let us dive deeper into ROUGE. It faced early criticisms regarding the stability and validity of human judgments, including potential limitations of reference texts from the dataset initially used for validation \cite{nenkova-passonneau-2004-evaluating}. The authors rebutted it later \cite{lin2004looking}, but still left some unanswered questions. However, others such as \citet{rankel2013decade} found that ROUGE-1, the variant most used, performed the weakest in a news summary data set, while R-4 was the strongest, and that combining ROUGE metrics improved performance, although with limited accuracy.\citet{cohan2016revisiting} observed that higher-order ROUGE metrics, particularly ROUGE-3, were better correlated with human judgments for scientific article summaries. \citet{peyrard2017learning} contradicted \citet{rankel2013decade}, finding that ROUGE-1 correlated best with human judgments, but on a relation extraction dataset. \citet{peyrard2019studying} showed that disagreement among ROUGE metrics increased for highly rated summaries, suggesting that human assessment remains essential for evaluating top-quality summaries. Another lexical metric, BLEU, has faced similar challenges, and numerous studies have sought to enhance its performance by modifying the calculation methodology \cite{post-2018-call}. 

In summary, lexical metrics provide a basic approach for evaluation, but fall short in capturing deeper meanings, which semantic similarity metrics try to address more efficiently.

\subsubsection{Semantic similarity metrics}
These metrics capture semantic similarity between generated and reference texts by representing them as numerical vectors (embeddings), capturing nuanced relationships across different linguistic levels. By applying distance measures to quantify the differences between the numerically represented texts, they leverage the characteristic that the closer the vector representations are, the more similar meaning of the texts \cite{harris1954distributional,mikolov2013distributed, de2021comparing}. 

The numeric representation of word meaning used in semantic metrics can be categorized according to whether they take into account the context of a token appearance or not. Context-independent word embeddings (for example, Word2Vec \cite{mikolov2013efficient}, GloVe \cite{pennington2014glove}) provide a one-word representation encoding for all its possible meanings. Contextual embeddings, on the other hand, adapt to specific contexts and can have multiple representations of the same word. An example is BERT \cite{devlin2019bert}, which generates contextual representations at a token level.

There are more complex numeric representations that go beyond tokens or words. Examples are Sentence-BERT \cite{reimers2019sentence}, which generates representations at the sentence level and fine-tunes BERT for semantically aligned sentences, and InferSent \cite{conneau2017supervised}, which uses a bidirectional LSTM trained on natural language inference datasets. These methods produce sentence representations that capture semantic structures beyond simple word embedding averaging.

To capture linguistic variations and fragmentations, the research field introduced various semantic similarity metrics based on the numeric representations presented above and respective distance measures. Some relevant examples are: \textbf{Word Mover's Distance (WMD)} \cite{kusner2015word}, \textbf{Sentence Mover's Distance} \cite{clark2019sentence}, \textbf{WISDM} \cite{botev2017word}, \textbf{BERTScore} \cite{Zhang2020BERTScore}, \textbf{BLEURT} \cite{sellam2020bleurt} and \textbf{BEM} \cite{bulian2022tomayto} (see Table \ref{tab:metrics_table}). In general, these metrics capture deeper semantic understanding, demonstrate superior performance in handling linguistic variations and linguistic fragmentations, and correlate better with human evaluations than lexical metrics \cite{sellam2020bleurt}.
 
However, they are not without limitations. For example, texts that are structurally similar to the reference but factually incorrect and often provide a poor account of syntax could lead to misleading scores \cite{chen2019evaluating}. For example, sentences with opposite meanings, such as with a negating particle, may be represented as similar. Although contextual embeddings like BERT capture some syntactic nuances, these often lack sufficient sensitivity to distinguish between syntactically similar yet semantically opposite statements.

Another general limitation of these metrics is their dependence on the embeddings used to calculate distances. The performance of the metric varies between different embeddings and can carry biased semantic information from its training data \cite{sun2022bertscore}. They also incur a higher computational cost, are slower than lexical metrics, and have limited generalization across unseen domains or languages, restricting their applicability.

Vector-based representations struggle to encode multiple meanings as well, having lower precision as semantic complexity increases. As a result, semantic similarity metrics perform well on simpler texts but fall short on more complex applications, and may not outperform lexical approaches in tasks requiring nuanced understanding of context and factual accuracy \cite{chen2019evaluating,sellam2020bleurt}. Difficulties with long-form text due to computational demands and loss of inter-sentence relationships have also been reported \cite{yeh-etal-2021-comprehensive}. These observations highlight the need for metrics that can handle complex responses with deeper contextual understanding.

\subsubsection{LLM-as-a-judge}
The LLM-as-a-judge metrics use generative models to evaluate free-text comparisons by responding to prompts and producing a score. Leveraging state-of-the-art LLMs can be a promising approach because they can be instructed to consider specific aspects of text through prompting, making them suitable for assessing complex text qualities. This type of metrics are developed for increasingly more powerful NLG systems and challenging datasets. The evaluation shifts from comparing many systems on one task to comparing a few LLMs on many datasets, with validations reported at the dataset level. These metrics can be used in various ways, including generating scores based on predefined criteria (scoring rubric), simulating human preferences through pairwise comparisons, or even operating without a reference. In this work, we focus on reference-based variations. Some examples in this category are: \textbf{BARTScore} \cite{yuan2021bartscore}, \textbf{GPT-Judge} \cite{lin2022truthfulqa}, \textbf{G-EVAL} \cite{liu2023g}, \textbf{GPT-Score} \cite{fu2024gptscore}, \textbf{Prometheus} \cite{kim2024prometheus}, and \textbf{AlpacaEval-LC} \cite{dubois2024length} (see Table \ref{tab:metrics_table}). 

Recent research has shown that LLM-based metrics, including the simple prompting of LLMs for an evaluation, often outperform traditional evaluation methods, correlating more strongly with human judgment \cite{chiang2023can,wang2023chatgpt,zheng2024judging}, despite being much slower and more expensive \cite{li2024leveraging}. Studies also suggest that additional measures are needed to improve the automatic calibration and alignment of LLM-based evaluators with human preferences \cite{liu2024calibrating,an-etal-2024-l}. This makes these metrics sensitive to variations in prompt design and data configurations \cite{li2024leveraging,wang2023chatgpt,kamalloo2023evaluating}, because different prompts can yield different scores, even if the aspect being assessed remains the same.  \cite{chiang2023can} analyze the components of LLM evaluation and find that asking LLMs to rationalize their ratings significantly improves correlation with human ratings \cite{fabbri2021summeval}. This underscores the importance of prompt design in LLM evaluations, as it enables LLMs to provide more informative ratings that align better with human judgments.

As a further caution it is worth noting that these metrics may favor responses based on surface-level attributes, such as verbosity, over substantive quality, potentially leading to high scores for linguistically refined but factually inaccurate responses \cite{chiesurin2023dangers}. Finally, there are also indications of inherent biases \cite{li2024leveraging}, including social biases inherited from the underlying models, model familiarity bias \cite{wataoka2024self}, and output length bias \cite{liu2024calibrating,an-etal-2024-l}. LLM-based evaluation methods have also shown inconsistencies compared to expert human evaluations \cite{stureborg2024large}.

\subsection{Correlation with human judgment}\label{human_correlation}

As shown in Table \ref{tab:metrics_table}, most evaluation metrics have been initially validated by assessing their correlation with human judgments on a specific task. Overall, the metrics designed to address application specific evaluation criteria have enabled NLP practitioners to develop and fine-tune their systems without human feedback.  However, validation methodologies vary significantly across metrics: they are tested on different tasks and datasets, use different correlation measures, apply varying statistical significance tests, and rely on human judgments of differing quality and quantity. This inconsistency makes direct comparisons challenging. Therefore, it is crucial to assess whether newer metrics truly outperform older ones. In the following, we present a meta-evaluation of metrics over time, analyzing their correlation with human judgments across diverse datasets, systems, and validation methodologies, beyond their initial assessments.

\subsubsection{Missing or inconsistent correlations}
Numerous studies have evaluated the correlation between metrics and human judgments, unfortunately often reporting either no correlation or inconsistent results.  \citet{liu2008correlation} found a low correlation between ROUGE and human evaluations of abstractive summaries of meetings, even when accounting for disfluencies and speaker information. \citet{graham2014testing} argued that claims of correlation with human assessments are insufficient to validate a metric and recommended statistical analyses, such as the Williams test, to confirm significance—an approach later adopted by BERTScore in its validation. 
 
In more recent years, \citet{deutsch2021statistical} used bootstrapping and permutation tests to calculate confidence intervals for correlations between metrics and human judgments in summarization tasks, observing wide confidence intervals, which highlights the high uncertainty in the reliability of AEMs. \citet{scalabrino2021assessing} analyzed correlations between human judgments and different metric combinations for generated code but were unable to find a satisfactory combination. \citet{liguori2023who} assessed a set of lexical metrics, including edit distance and exact match, to evaluate harmful code production, finding that ROUGE-4 and BLEU-4, often used for that use case, failed to estimate semantic correctness, whereas exact match and edit distance exhibited the highest correlation with human evaluations. More recently, \citet{li2024leveraging} compiled a taxonomy of LLM-based NLG evaluators and compared their performance with older metrics across various tasks, observing that LLM-as-a-judge metrics correlated better with human judgments, yet correlations remained low to moderate and exhibited high variance across methods. 

\subsubsection{Influencing factors}
Some studies have attempted to explain these inconsistencies by identifying specific influencing factors. One widely discussed factor is that correlation scores vary depending on the reference used, meaning that results can differ across datasets.  \citet{bhandari2020metrics} found that metrics correlate better for easier-to-summarize documents, with performance deteriorating as summaries become more abstractive. Similarly, \citet{bhandari2020re} analyzed correlations between lexical and semantic metrics and human assessments for several summarization systems and tasks, showing that different metrics correlate better depending on the dataset.  \citet{moramarco2022human} compared metrics for medical note generation, including edit distance, lexical, and semantic metrics, and found that results varied depending on the reference text. \citet{zhang2024benchmarking} examined human ratings of GPT-3 summaries for well-known news datasets across several dimensions and found that correlations fluctuated significantly depending on the dataset and reference summaries. In particular, substituting reference summaries with those written by freelance writers improved the correlation with ROUGE-L for faithfulness, emphasizing the impact of reference summary quality when evaluating reference-based metrics.

\subsubsection{Quality of evaluated system} 
Another influencing factor is the quality of the system under evaluation.  \cite{novikova2017we} found that lexical metrics exhibited a poor correlation with human judgement for general NLG tasks, particularly struggling to differentiate between medium- and high-quality outputs. However, AEMs were more effective in identifying low-performing systems. \citet{mathur2020tangled} compared the correlation between lexical and semantic metrics, highlighting the sensitivity of Pearson’s correlation to sample size. They found that metric correlations weakened for top-performing systems and similarly performing models, with BLEU’s performance deteriorating when high-performing systems were removed, exposing the fragility of score aggregation.

\subsubsection{Expertise of human annotators}
One more factor affecting correlation is the expertise of human annotators.  \citet{reiter2009investigation} compared lexical metrics for a weather forecast NLG task and found inconsistent results between expert and non-expert judgments, with most metrics favoring non-experts. \citet{bavaresco2024llms} compiled a collection of 20 NLP datasets with human annotations and compared them with results from multiple open and closed LLMs. They found that Spearman’s correlation was consistently higher when comparing models against non-expert judgments rather than expert assessments, with the highest correlations observed in the verbosity dimension. Their analysis also showed that no single model consistently outperformed others in all evaluation categories and chain-of-thought (CoT) prompting did not consistently improve agreement with human assessments.

\subsubsection{System vs segment level correlations}
A key observation across studies is that the performance of the metrics differs depending on the level of analysis, system vs segment level correlations. \citet{novikova2017we} confirmed that metrics perform better at the system level than at the sentence level. \citet{bhandari2020re} further demonstrated that these metrics struggle to quantify system improvements and that document-level comparisons can yield different conclusions from system-level comparisons.

\subsubsection{Evaluation aspect}
Finally, some studies have found that metric correlations vary depending on the specific evaluation aspect being considered. \citet{stent2005evaluating} concluded that automatic metrics were better suited for evaluating adequacy than fluency. In contrast, \citet{reiter2009investigation} found that most metrics were more strongly correlated with fluency than with accuracy. Similarly, \citet{stent2005evaluating} found no significant correlation between human judgments of adequacy and fluency for generated paraphrases with controlled syntactic variations. \citet{fabbri2021summeval} introduced SummEVAL, where re-annotated summary datasets are used to compare correlations between metrics and evaluation aspects (coherence, consistency, fluency, and relevance), finding substantial discrepancies across them.

Taken together, these findings underscore the inconsistency - and consequently the inconclusiveness - of metric correlations with human judgments. Greater attention must be paid to validation methodologies to clearly determine what aspects AEMs reliably capture and how they should be used, ultimately improving their effectiveness.

\subsection{Evaluation in real scenarios: RAG systems}\label{rag}

The evaluation metrics discussed so far aim to quantify the similarity between a model output and a reference text. However, real-world NLG applications introduce additional performance criteria which cannot be accounted for in terms of text similarity. An example of such an application is RAG, which complements NLG with information retrieval mechanisms.  The goal of RAG is to minimize the tendency of generative LLMs to produce factually incorrect information (i.e. “hallucinate”) and to allow them to use data which have not been part of their training set. In a typical RAG setup, a user query is first used to find relevant documents, and then a generative LLM is asked to respond to the query given the documents. Importantly, the expectation of a RAG system is that its output is grounded in the provided context, e.g., it is based on retrieved documents and not on the parametric knowledge of the underlying LLM \cite{jacoviFACTSGroundingLeaderboard2025}.

The specifics and inherent complexity of RAG require considering multiple aspects of its evaluation \cite{chen2023RAG}. Some proposed criteria include faithfulness (the degree to which the text is grounded in the provided context), answer relevance (whether the output actually addresses the user query), completeness (whether all the relevant information provided by the context is utilized), noise robustness (whether the system can ignore irrelevant information), and information integration (the ability to combine information from multiple sources). The dominant approach to implementing these criteria is the ‘LLM-as-a-judge’, i.e. by prompting an external LLM. For example, the \textbf{Tonic} evaluation framework \cite{TonicValidate2023} computes RAG faithfulness by extracting a list of claims which can be inferred from the output text and checking the proportion of them which can be attributed to the provided context. In a similar vein, the \textbf{RAGAS} framework \cite{es2024ragas} estimates answer relevance by using an LLM to generate a list of probable user queries which the answer could be responding to and calculate their similarity to the real query.  However, to our point, the initial RAGAS metrics validation has used a statistically insignificant number of human annotators (two) to calculate correlation with human judgment, and we expected future studies to challenge those results. 

Recent models like IntellBot \cite{arikkat2024intellbot} and QuIM-RAG \cite{saha2024advancing} do incorporate RAGAS for evaluation, but the over reliance on complicated LLM prompts and the lack of compelling evidence in support of the validity of these metrics undermines their credibility. It illustrates the current landscape of metric usage that other RAG system evaluations still rely heavily on traditional metrics, in spite of its specific evaluation needs. The original RAG paper \cite{guu2020retrieval} used exact matching (EM) with BLEU and ROUGE, while later studies \cite{hsu2021answer,giglou2024scholarly} advocate a hybrid approach introducing semantic metrics like BERTScore and BLEURT. 

As presented, even modern real world applications such as RAG we see the same shortcomings: lack of statistical significance, inconsistency with human judgement and unfounded usage of metrics without considering the specifics of the use case and what the metrics are measuring.

\section{Threats to validity}\label{threats_validity}
The many inconsistencies revealed by attempts to reproduce the correlations between metric scores and human annotations could also be amplified by issues that threaten the validity of the correlations themselves.

Firstly, the quality of human-written references can significantly affect the reliability of metrics \cite{gehrmann2023repairing,kamalloo2023evaluating}. Ensuring reference texts are of high quality is crucial, yet challenging. Furthermore, annotators' agreement can reflect inconsistencies even within the same annotator across different evaluations \cite{abercrombie2023temporal}. Even when references are assumed to represent a "gold standard", accurately measuring similarity between generated texts and references remains difficult.

Secondly, human judgment itself introduces validity threats. Recent studies indicate that human evaluators occasionally prefer model-generated texts to human-written ones \cite{gehrmann2023repairing,sottana2023evaluation}. For example, \citet{goyal2022news} compare metric correlations with human judgments in the era of LLMs and find that human evaluators prefer GPT3 summaries, while reference-based metrics show an inverse correlation, concluding that current automatic metrics are inadequate for evaluating new generative models. This aspect of human evaluation requires further investigation.

\section{Conclusion}\label{conclusion}
This paper examines NLG AEMs, analyzing their methodologies, strengths, limitations, initial validation approaches, and alignment with human judgment. Despite ongoing research, no single metric has been definitive and NLG systems are still validated using varied metrics without fully considering the implications, which we highlight in the case of RAG.

Each type of metric has specific strengths: lexical metrics suit fact-based short texts; semantic similarity metrics work well for paraphrases; and LLM-based metrics are better for complex texts when computational resources allow. They all have limitations too: lexical metrics rely on surface similarity, semantic similarity metrics may overlook syntax, and LLM-based metrics are costly and sensitive to prompt formulation.

Broader concerns exist regarding metric validity. Due to differences in validation methods, direct comparisons between metrics are often unreliable and reported correlations with human judgment are inconsistent. Furthermore, we find problems with human annotations themselves, which call into question the validity of the correlations.

Our key insight is that metrics, at best, highlight specific dimensions of text quality, since they are still widely used, but they do not provide absolute performance measures. The pursuit of a “perfect metric” is thus misguided; instead, we advocate for a task-specific approach, to avoid oversimplifying text quality assessment, and for further investigations on which aspects of text quality are in fact evaluated by each metric. Future research should also seek to refine metric selection, improve validation methodologies, and establish best practices to enhance reliability and alignment with human preference.

%
%
%
\bibliographystyle{acl_natbib}
\bibliography{references_submission}
%

\end{document}